\documentclass[runningheads,a4paper]{llncs}

\usepackage[utf8]{inputenc} 

\usepackage{amsmath}
\usepackage{amssymb}
\usepackage{graphicx}

\usepackage{url}
\newcommand{\keywords}[1]{\par\addvspace\baselineskip
\noindent\keywordname\enspace\ignorespaces#1}

\usepackage[colorlinks]{hyperref}
\usepackage[]{longtable}
\usepackage[]{booktabs}
\usepackage{subfigure}
\usepackage{mdframed}
\usepackage{etoolbox}
\usepackage{array}

\definecolor{Gray}{gray}{0.8}

\newcolumntype{g}{>{\columncolor{Gray}}c}
\newcolumntype{f}{>{\centering\arraybackslash}p{16mm}}
\newcolumntype{h}{>{\centering\arraybackslash}p{9.2mm}}

\begin{document}

\mainmatter  

\title{LSA64: An Argentinian Sign Language Dataset}

\titlerunning{LSA64: An Argentinian Sign Language Dataset}

\author{Franco Ronchetti* \inst{1} \and Facundo Quiroga* \inst{1} \and César Estrebou \inst{1} \and Laura Lanzarini \inst{1} \and Alejandro Rosete \inst{2} }
\authorrunning{F. Ronchetti et al.}

\def\aa{\}}
\def\ab{\{}

 \institute{
 Instituto de Investigación en Informática LIDI, Facultad de informática,\\
   Universidad Nacional de La Plata
  \\ \email{\ab fronchetti,fquiroga,cesarest,laural\aa @lidi.unlp.edu.ar}
 \and
   Instituto Superior Politécnico Jose Antonio Echeverría
  \\ \email{\ab rosete\aa @ceis.cujae.edu.cu}   \footnotetext{* Contributed equally }
 }

\toctitle{}
\tocauthor{}
\maketitle

\begin{abstract}

Automatic sign language recognition is a research area that encompasses human-computer interaction, computer vision and machine learning. Robust automatic recognition of sign language could assist in the translation process and the integration of hearing-impaired people, as well as the teaching of sign language to the hearing population.

Sign languages differ significantly in different countries and even regions, and their syntax and semantics are different as well from those of written languages. While the techniques for automatic sign language recognition are mostly the same for different languages, training a recognition system for a new language requires having an entire dataset for that language.

This paper presents a dataset of 64 signs from the Argentinian Sign Language (LSA). The dataset, called LSA64, contains 3200 videos of 64 different LSA signs recorded by 10 subjects, and is a first step towards building a comprehensive research-level dataset of Argentinian signs, specifically tailored to sign language recognition or other machine learning tasks. The subjects that performed the signs wore colored gloves to ease the hand tracking and segmentation steps, allowing experiments on the dataset to focus specifically on the recognition of signs.

We also present a pre-processed version of the dataset, from which we computed statistics of movement, position and handshape of the signs.


\end{abstract}

\section{Introduction}

Sign language (SL) recognition is a complex multidisciplinary problem. It bears many similarities to speech recognition, but presents some additional difficulties \cite{von2008recent}:

\begin{enumerate}
  \item There is little formal standardization in most sign languages, even within a region.
  \item Sign language specification languages themselves are not well standardized, and there is no consensus on which type of specification is more appropriate.
  \item Signs are intrinsically multimodal: they are formed by a combination of hand shapes, positions, movements, body pose, face expression, and lip movements. In contrast, speech recognition usually requires only sound input.
  \item Creating datasets for training sign language recognition systems requires being able to capture and model all of these signals.
  \item There are relatively few sign language users with respect to the general population, and therefore finding expert subjects to record signs is more difficult.
\end{enumerate}

For these reasons, it is generally considered that we are still far from robust sign language recognition systems.

There are numerous publications dealing with the automatic recognition of sign languages, and \cite{Cooper2011a,von2008recent} present reviews of the state of the art in sign language recognition. The full task of recognizing a sign language involves a multi-step process. In the context of video-based recognition, and considering only manual information, this process can be simplified as:

\begin{enumerate}
  \item Tracking and segmenting the hands of the interpreter in every frame of the video.
  \item Recognizing the shapes of the hands, the movements they made and their positions.
  \item Recognizing the sign as a syntactic entity (a visual word).
  \item Assigning semantics to a sequence of signs (a visual sentence).
  \item Translating the semantics of the signs to the written language.
\end{enumerate}

These tasks are mostly independent from each other, and involve different techniques. Tracking, segmentation and modeling of the hand are mostly signal processing and 3D modeling tasks, while assigning semantics to a sequence of signs and translating from sign to written languages are more related to natural language processing.

\subsection{Datasets}

There are many datasets for sign language recognition. We distinguish three types, depending on the problem they target: handshape recognition, sign recognition or sentence recognition. Each type of dataset presents a greater challenge than the previous one, and allows experiments with more steps of the recognition pipeline.

Table \ref{tab:dataset} presents the most prominent video-based, research-level datasets for recognition. Since the dataset described in this paper focuses on sign-level recognition, we only list sign and sentence-level datasets \footnote{A more detailed reference about sign language datasets can be found at \url{http://facundoq.github.io/guides/sign_language_datasets/slr}}.

\begin{table}
\centering
\caption{Recognition-oriented, video based, sign language datasets used in recent papers.}
\label{tab:dataset}
\begin{tabular}[c]{cccccccccc}
  \toprule
  Name  & Classes & Subjects & Samples &  Language
  level & Availability
  \\
  \midrule
    DGS Kinect 40 \cite{Cooper2011a} & 40 & 15 & 3000 & Word & Contact Author
  \\\addlinespace
  DGS RWTH-Weather \cite{von2008recent} & 1200 & 9 & 45760 & Sentence & Public Website
  \\\addlinespace
 DGS SIGNUM \cite{von2008recent} &  450 & 25 & 33210 & Sentence &  Contact Author
\\\addlinespace
 GSL 20 \cite{Cooper2011a} &  20 & 6 & 840 & Word &  Contact Author
\\\addlinespace
 Boston ASL LVD \cite{neidle2012challenges} &  3300+ & 6 & 9800 & Word & Public Website
\\\addlinespace
 PSL Kinect 30 \cite{kapuscinski2015recognition}  & 30 & 1 & 300 & Word & Public Website
\\\addlinespace
 PSL ToF 84 \cite{kapuscinski2015recognition}  & 84 & 1 & 1680 & Word & Public Website
\\\addlinespace
 \bottomrule
\end{tabular}
\end{table}

In general, the datasets that are video-based must rely on skin color tracking and segmentation, and are therefore are not robust for background variations or interpreter clothes, as well as hand-hand or hand-face occlusions \cite{Roussos2010a}. Usually, to perform features extraction these datasets require the addition of morphological information as a subsequent step to the color filtering to identify the position and shape of the hand, which can be extracted using depth cameras or other sensors, but these limit the application of the methods with respect to using normal video cameras, given the availability of each type of devices.

The largest sign language dataset available (in terms of number of classes), the \textit{American Sign Language Lexicon Video Dataset} (ASLLVD) \cite{neidle2012challenges}, contains more than 3300 signs from the American Sign Language, but near-perfect hand tracking and segmentation on this dataset is difficult \cite{neidle2012challenges}, making it hard to use it to evaluate a sign recognizer that focuses on the syntactic and semantic recognition. The situation is similar for the SIGNUM and RWTH-Weather datasets. Moreover, the dataset ASLLVD has only 6 subjects, and an average of 3 samples per class.

\subsection{Argentinian Sign Language Dataset (LSA64)}

Sign languages are different in each region of the world, and each has their own lexicon and group of signs. Thus, sign language recognition is a problem that needs to be tackled differently in each region, since new movements or handshapes or combinations thereof require new training data, and possibly involve new challenges that were not considered before \cite{von2008recent,Cooper2011a}.

To the best of our knowledge, there are no available datasets for the Argentinian Sign Language (LSA). There are only a few dictionaries that focus on teaching the language. Since they were recorded with this aim in mind, they have only one sample for each sign, low image quality, and poor annotations, thus making them unsuitable for training automatic recognition systems. There is need for a research-level dataset that represents the group of signs used in LSA.

This paper presents a sign dataset called LSA64. The dataset consists of 64 signs from the LSA, and was recorded with normal RGB cameras. It is publicly available\footnote{The dataset and relevant information can be found at \url{https://facundoq.github.io/datasets/lsa64}}, and we also provide a preprocessed version of the dataset to facilitate experiments and reproducibility.

The subjects wore colored gloves for the recording (single colored gloves, with different colors for each hand). This methodology allows researchers to easily bypass the tracking and segmentation steps, and focus on the subsequent steps of the recognition \cite{wang2009real}.

While the dataset has less classes than ASLLVD, RWTH-PHOENIX-Weather or SIGNUM, it has more samples and subjects than many other datasets (Table \ref{tab:dataset}), and it is publicly available with a preprocessed version.

The document is organized as follows: Section 2 describes the LSA64 dataset and the recording conditions. Section 3 presents statistics and information of the signs recorded, to aid in the understanding of the dataset. Section 4 details an experiment carried out to establish a baseline on this dataset, and Section 5 presents the general conclusions.

\section{Dataset}

\begin{figure}
\centering
\includegraphics[width=0.325\textwidth]{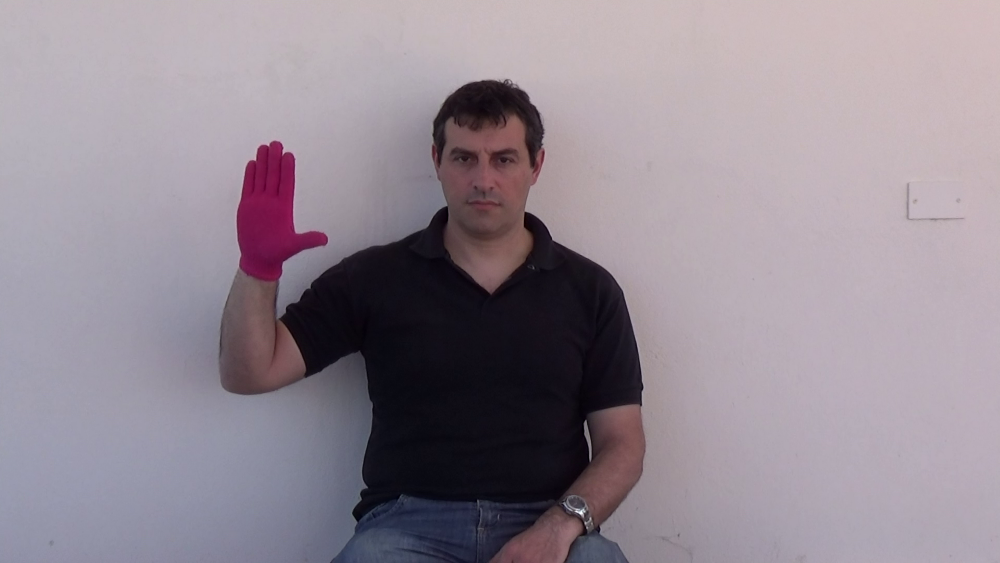}
\includegraphics[width=0.325\textwidth]{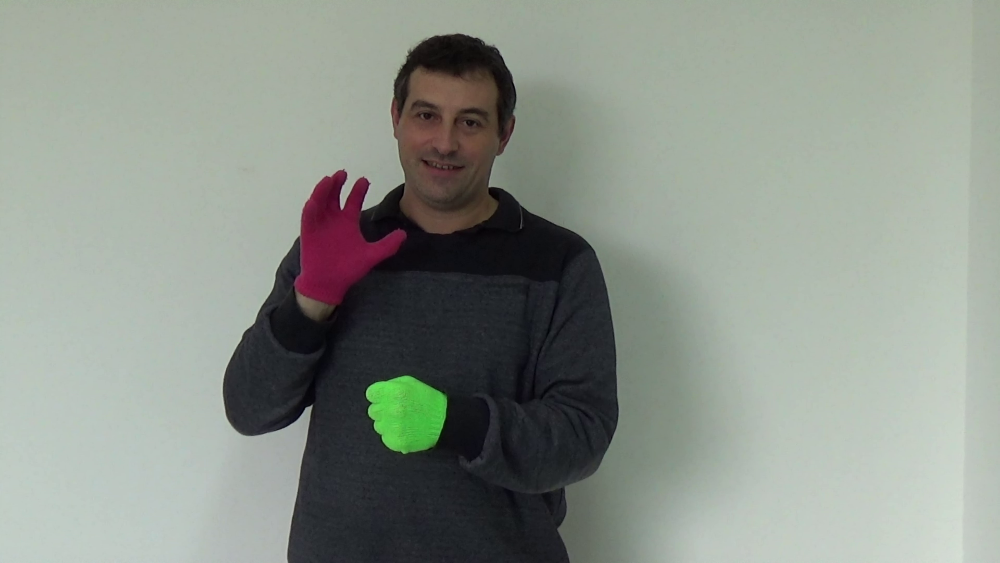}
\includegraphics[width=0.325\textwidth]{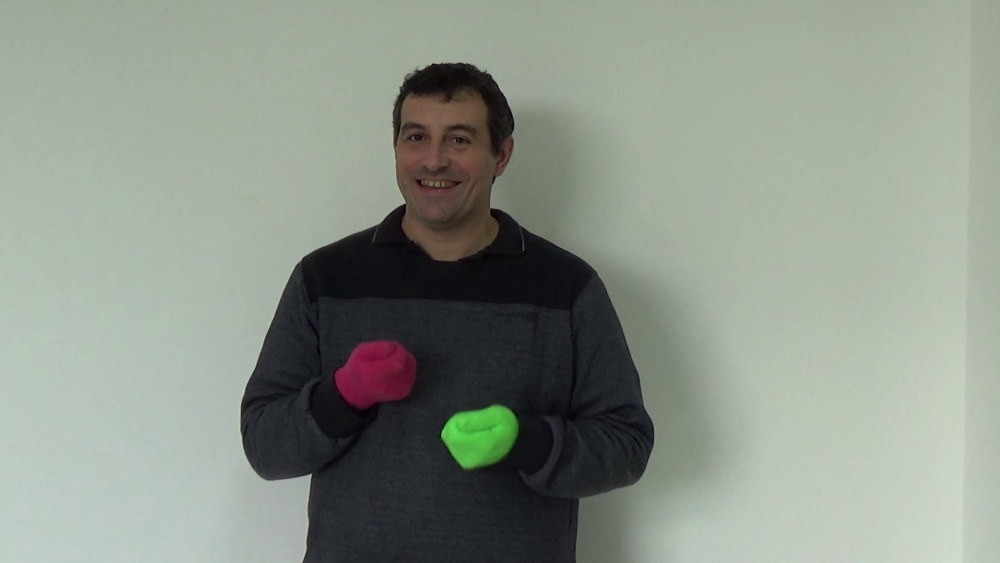}
\\ \vspace{0.5mm}
\includegraphics[width=0.325\textwidth]{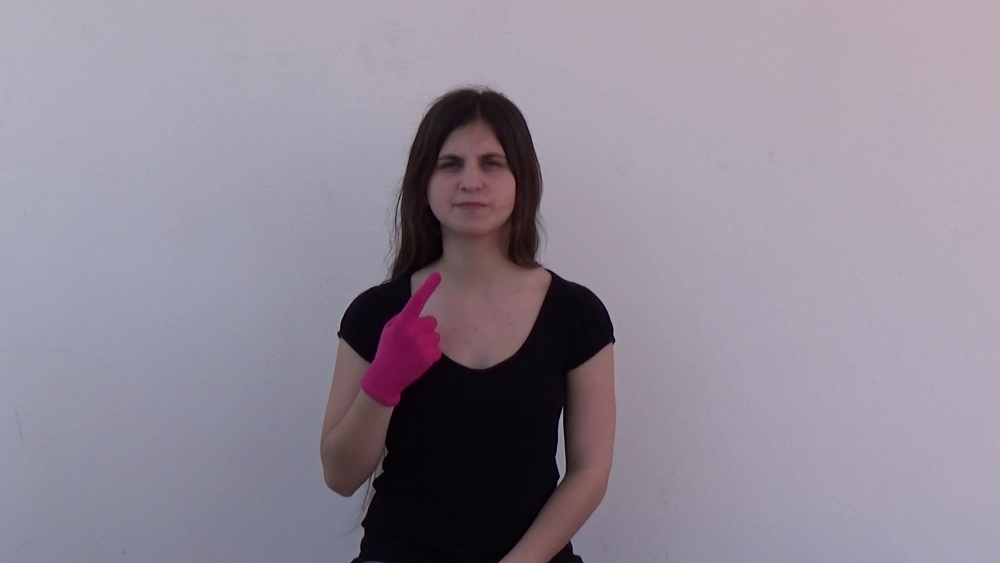}
\includegraphics[width=0.325\textwidth]{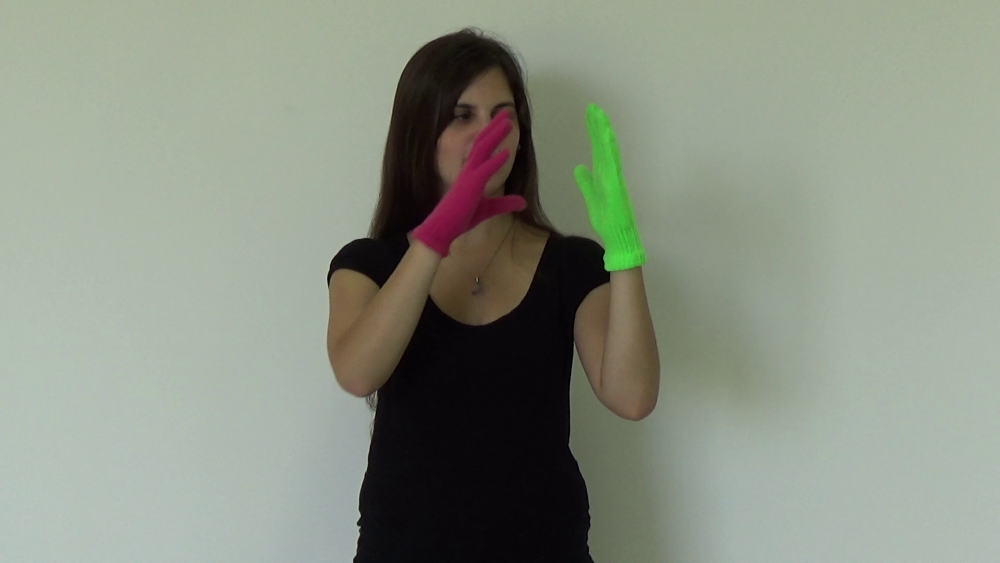}
\includegraphics[width=0.325\textwidth]{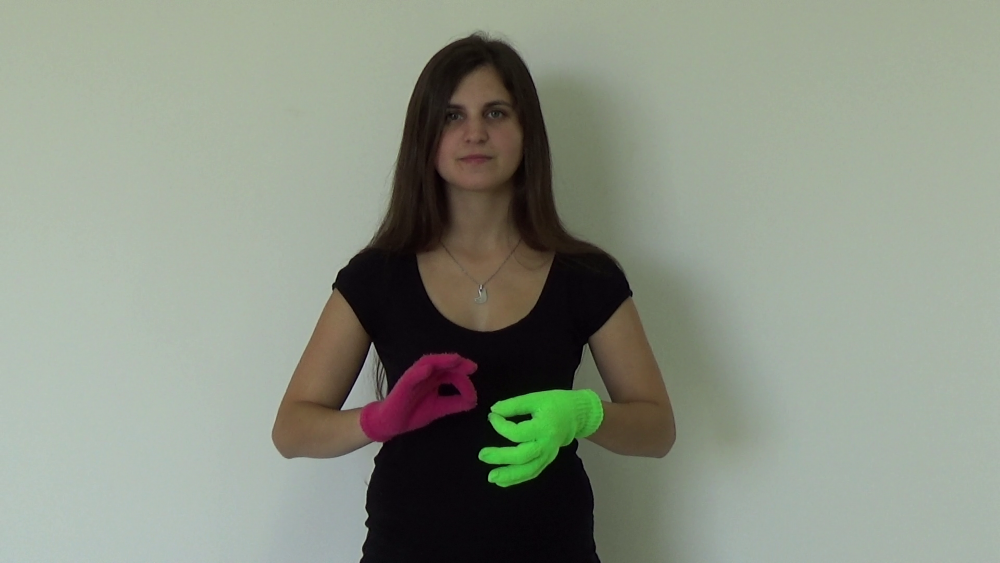}
\caption{Snapshots of six different signs of the LSA64 database. There are overlaps in positions and handshapes. The images on the left are from the first set of recordings.}
\label{fig:lsa64}
\end{figure}

The sign database for the Argentinian Sign Language, created with the goal of producing a dictionary for LSA and training an automatic sign recognizer, includes 3200 videos where 10 non-expert subjects executed 5 repetitions of 64 different types of signs. Signs were selected among the most common in the LSA lexicon, and include both verbs and nouns. Some examples can be seen in Figure \ref{fig:lsa64}.

\subsection{Recording}

The database was recorded in two sets. In the first one, 23 one-handed signs were recorded. The second added 41 signs, 22 two-handed and 19 one-handed. The final dataset then contains 42 one-handed signs and 22 two-handed ones.

The first recording was done in an outdoors environment, with natural lightning, while the second took place  indoors, with artificial lightning (Figure \ref{fig:lsa64}). Subject 10 from the first recordings was unavailable for the second set of recordings, and was replaced by another subject. This change in no way diminishes the utility of the dataset, since the set of classes recorded in the first session is disjoint from the ones recorded in the second session.

In both sets of recordings, subjects wore black clothes and executed the signs standing or sitting, with a white wall as a background. To simplify the problem of hand segmentation within an image, subjects wore fluorescent-colored gloves. These substantially simplify the problem of recognizing the position of the hand and performing its segmentation, and remove all issues associated to skin color variations, while fully retaining the difficulty of recognizing the handshape. Additionally, each sign was executed imposing few constraints on the subjects to increase diversity and realism in the database. The camera employed was a Sony HDR-CX240. The tripod was placed 2m away from the wall at a height of 1.5m.



In the following subsections we show statistics and information of the signs to better understand the nature and challenges of the dataset. These statistics show that signs in this dataset possess significant overlap in terms of types of movements, initial and final positions and handshapes, producing non-trivial experiment settings to test new sign language recognition models. All the information has been computed from the pre-processed version of the dataset described in Section \ref{sec:preprocessing}.

\subsection{Handshapes}

\begin{figure}
\centering
\includegraphics[width=0.9\textwidth]{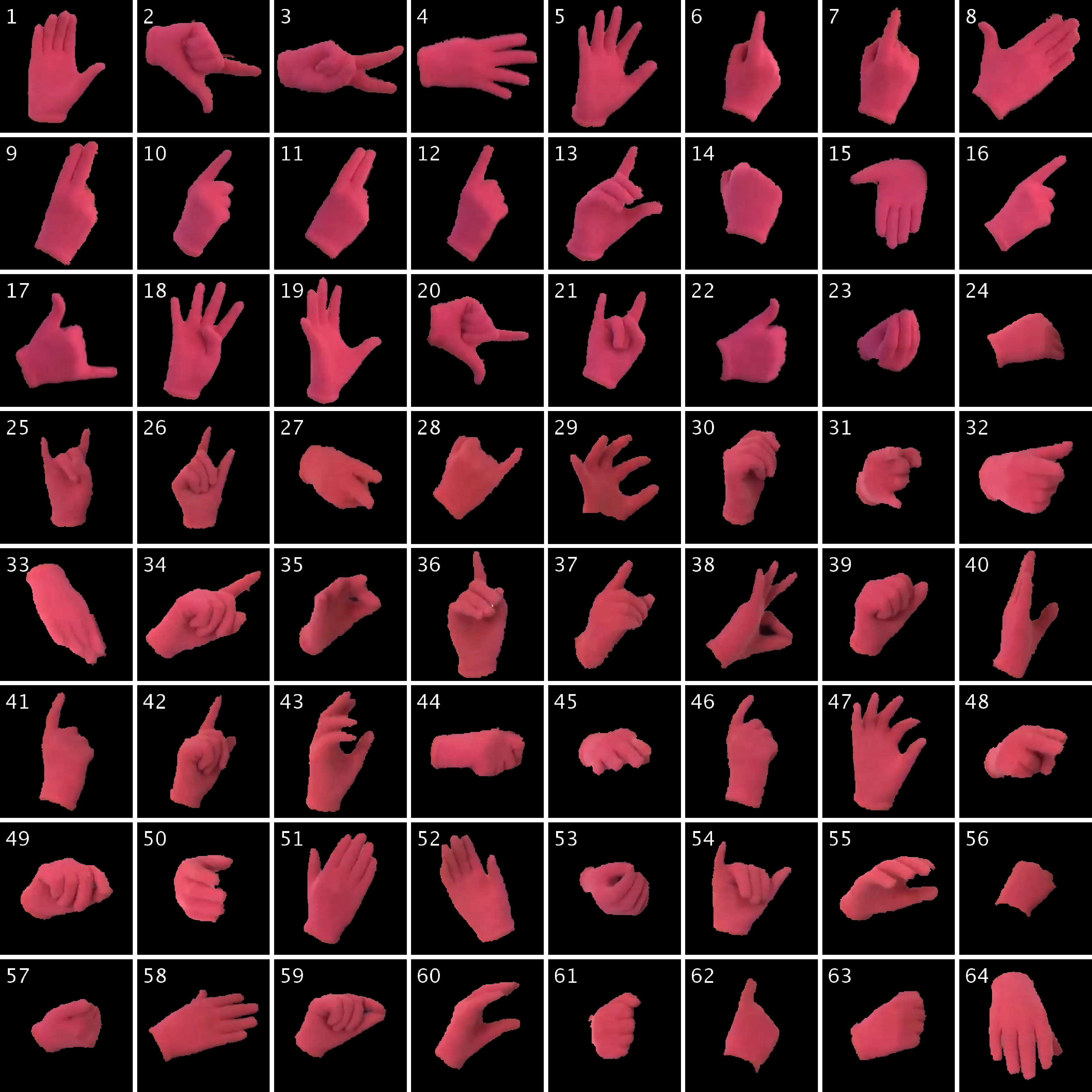}
\caption{Images of segmented hands as captured in the LSA64 database. Each image shows the initial handshape of the right hand for each sign in the dataset.}
\label{fig:lsa64handshapes_right}
\end{figure}

\begin{figure}
	\centering
	\includegraphics[width=0.9\textwidth]{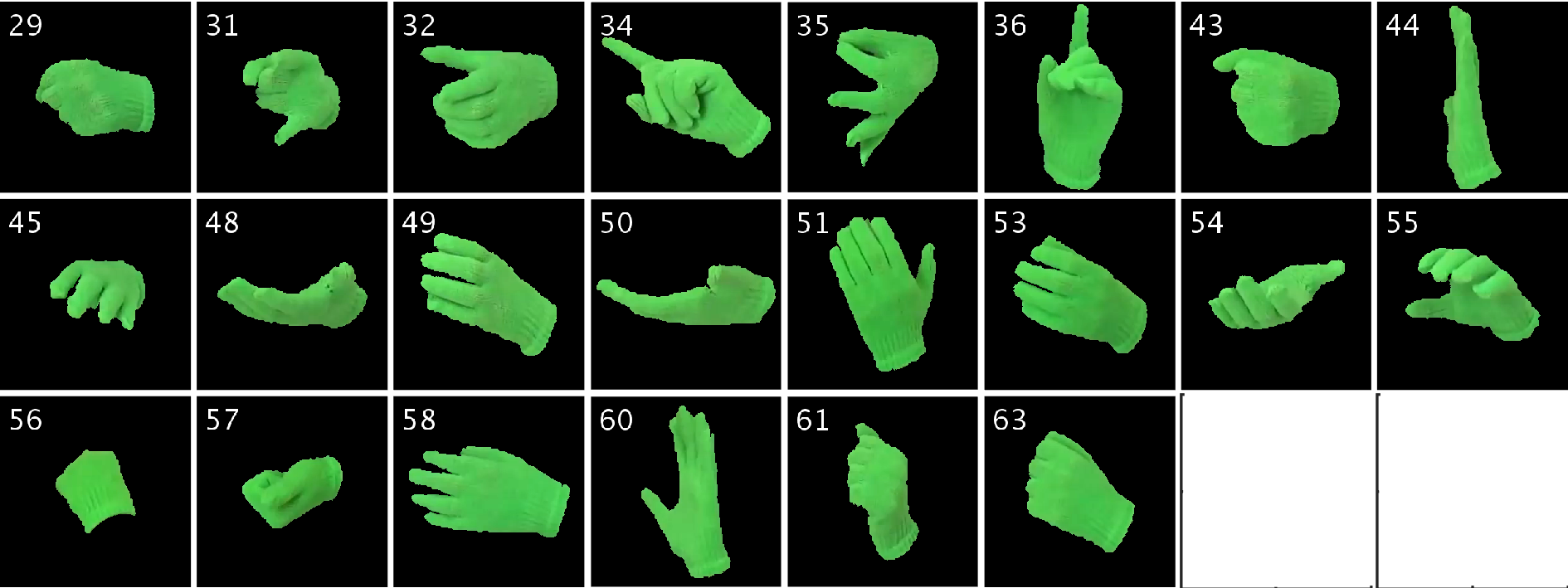}
	\caption{Images of segmented hands as captured in the LSA64 database. Each image shows the initial handshape of the left hand for the two-handed signs of the dataset.}
	\label{fig:lsa64handshapes_left}
\end{figure}

In Figures \ref{fig:lsa64handshapes_right} and \ref{fig:lsa64handshapes_left} we can observe the different handshapes of the right and left hand respectively for each class of sign. There is plenty of repetition between handshapes, although their 2D projection may be different depending on the rotation of the hand.

\subsection{Positions}

\begin{figure}
\centering

\subfigure[]{
\includegraphics[width=0.22\textwidth]{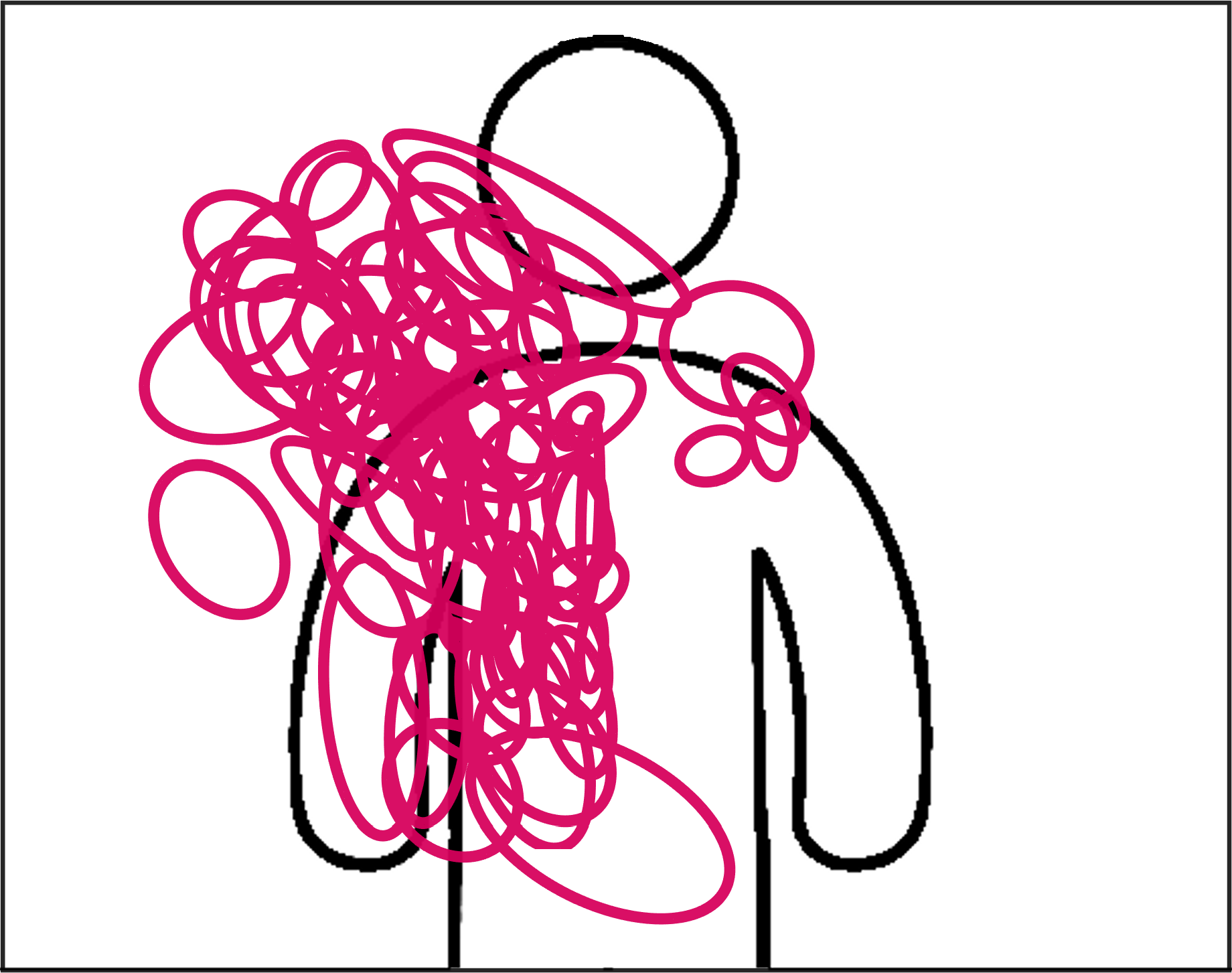} 
\label{fig:lsa64positionrightinitial}
}
\subfigure[]{
\includegraphics[width=0.22\textwidth]{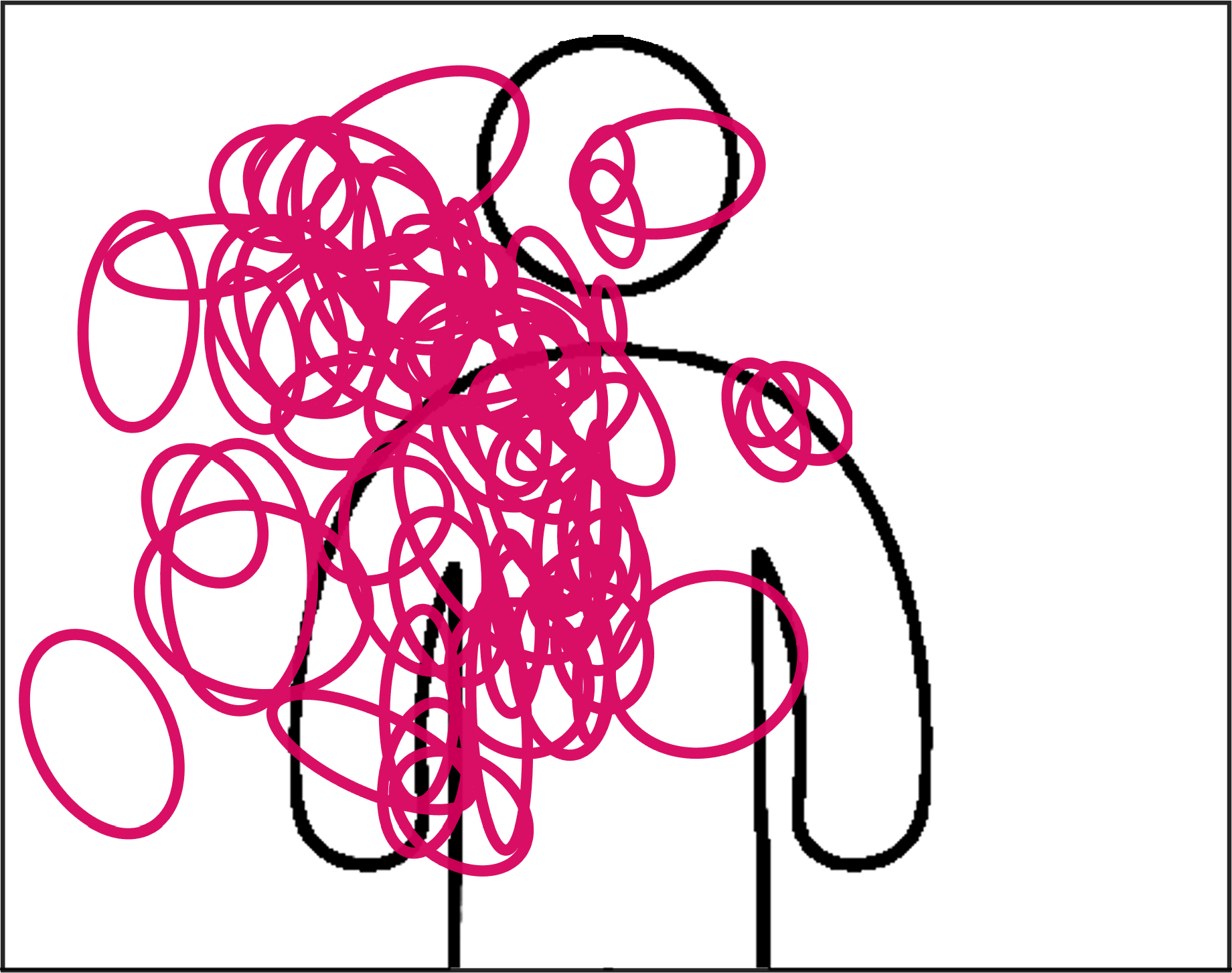}
\label{fig:lsa64positionrightfinal}
}
\subfigure[]{
	\includegraphics[width=0.22\textwidth]{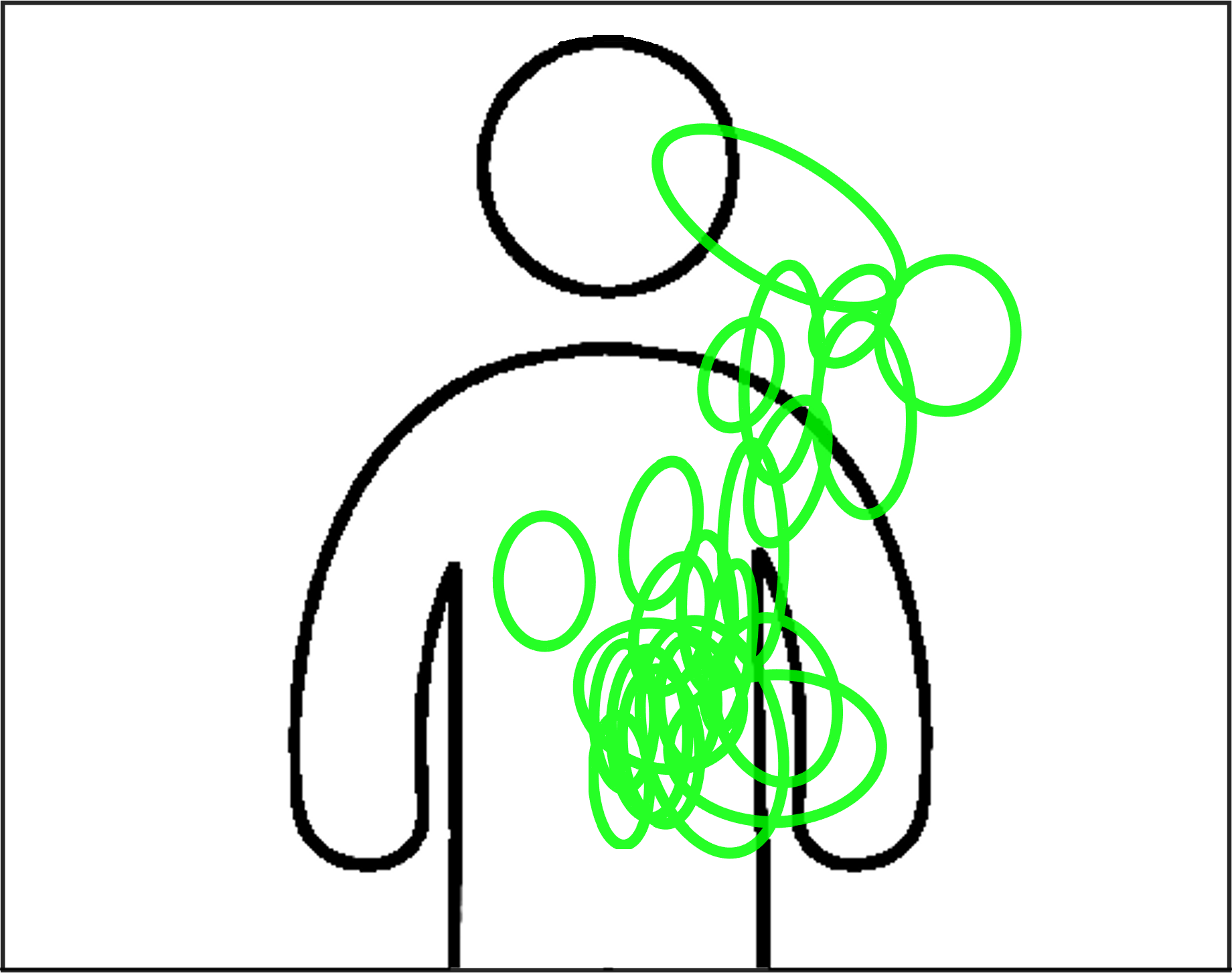} 
	\label{fig:lsa64positionleftinitial}
}
\subfigure[]{
\includegraphics[width=0.22\textwidth]{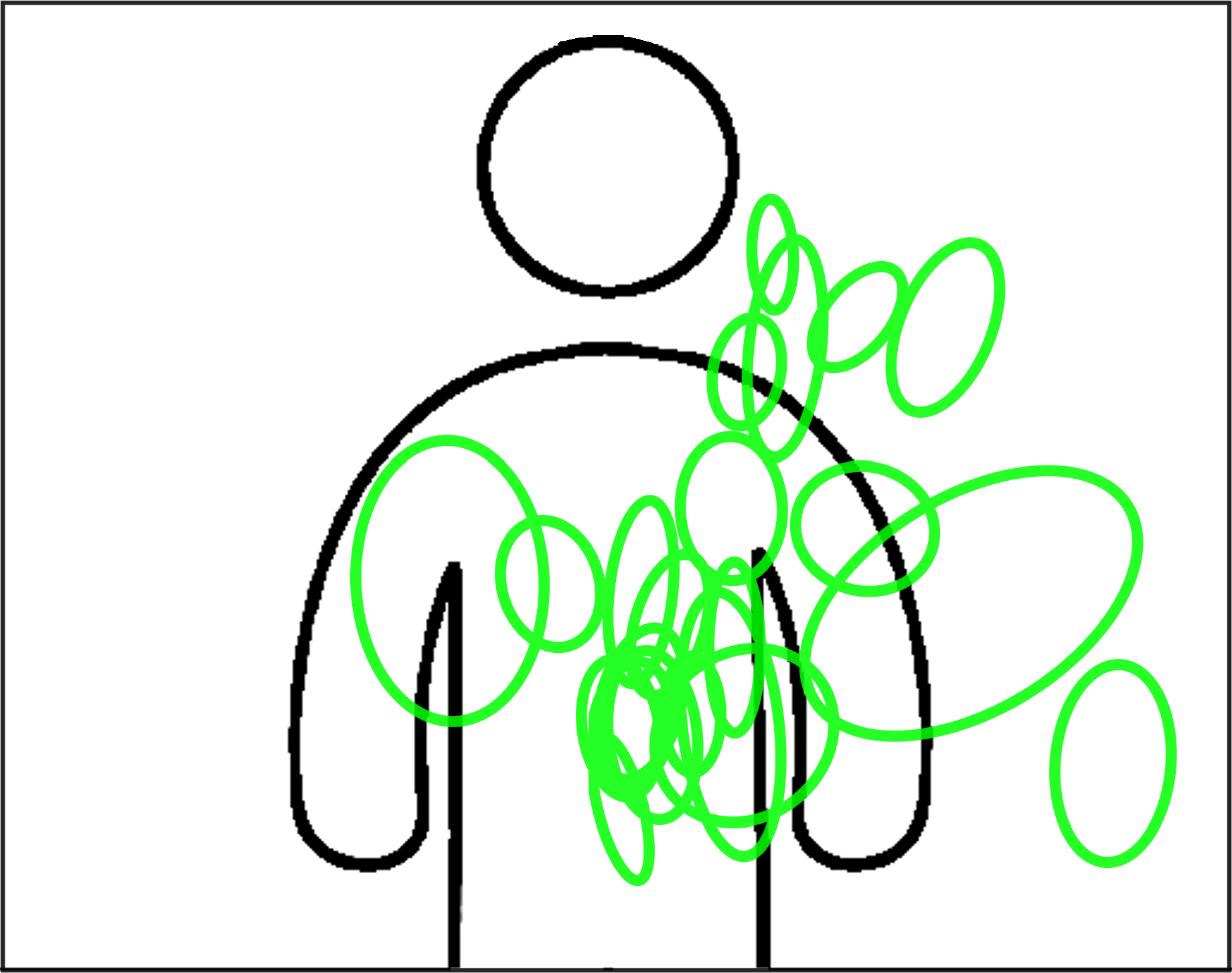}
\label{fig:lsa64positionleftfinal}
}
\caption{Means for the initial and final positions of the right hand for each sign (a and b), and also for the left hand (c and d). Circles around means represent the covariance of the samples.}
\label{fig:lsa64position}
\end{figure}

Figure \ref{fig:lsa64position} presents the mean initial and final positions for each hand, along with the covariance. While a few signs can be identified by their positions, they overlap significantly in most cases.

\subsection{Trajectories}

\begin{figure}
\centering
\includegraphics[width=1\textwidth]{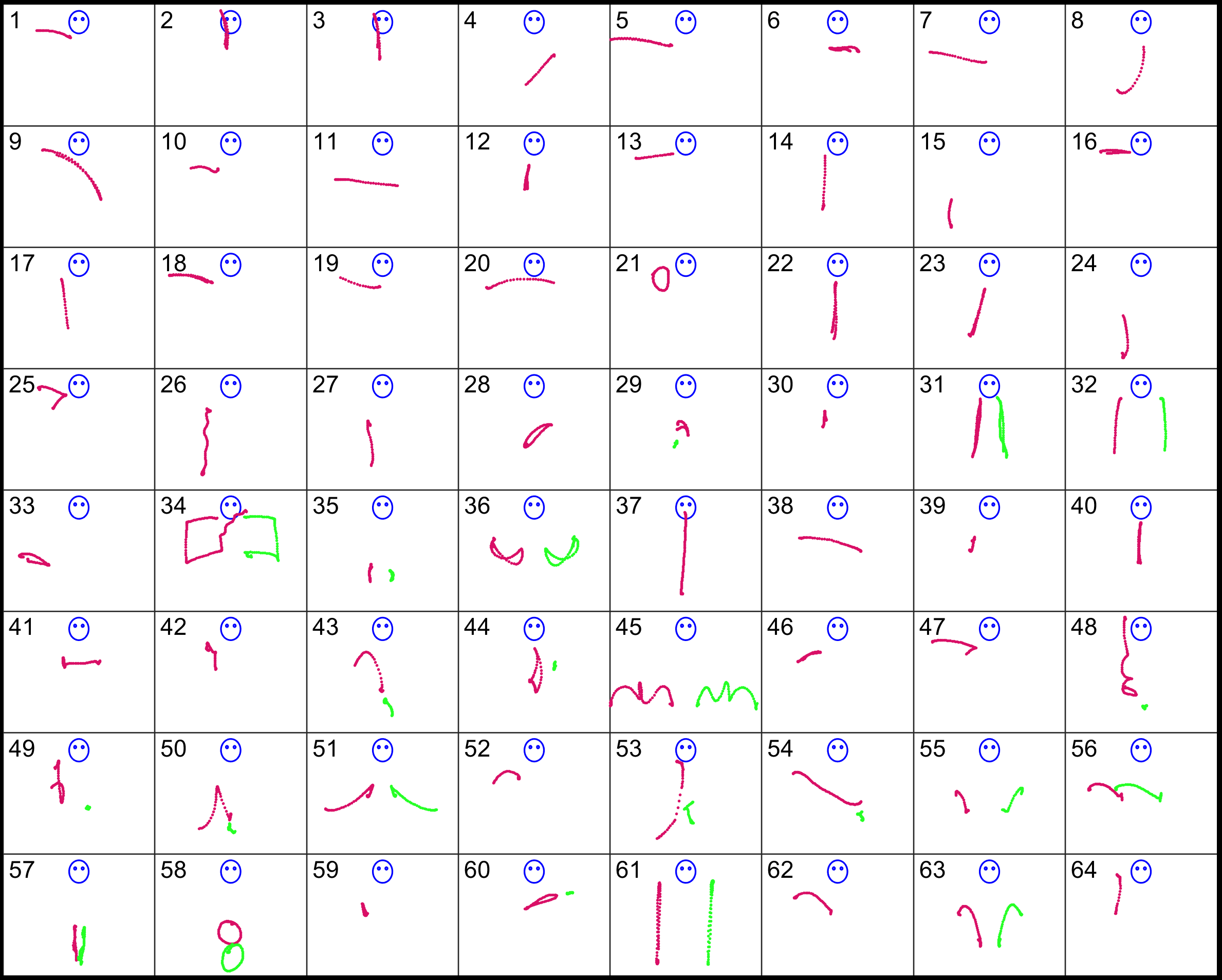}
\caption{Sample trayectories for each sign in LSA64. The left-hand trajectory is shown in light green, the right-hand one in red, and the head position as a blue circle. }
\label{fig:lsa64movement}
\end{figure}

Figure \ref{fig:lsa64movement} shows sample trajectories of each sign, as performed by subject 2. There is much overlap in movements for both one-handed (for example, signs 1, 5, 7, 13 and 19) and two-handed signs (for example, signs 31, 32 and 61).

\subsection{Amount of movement}
\begin{figure}
\centering
\includegraphics[width=0.9\textwidth]{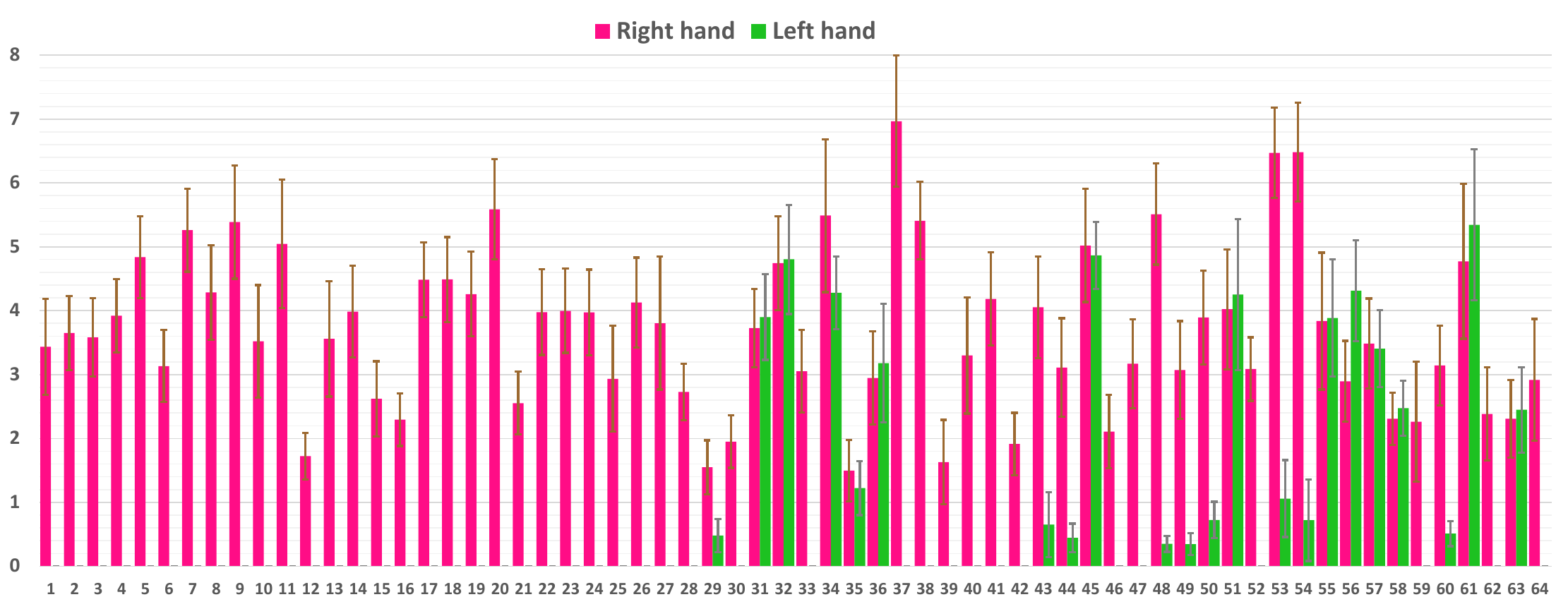}
\caption{Amount of movement for each class of sign. Red bars show the amount of movement of the right hand, and green bars the movement of the left hand (in two-handed signs).}
\label{fig:lsa64_amount_movement}
\end{figure}

Figure \ref{fig:lsa64_amount_movement} shows the amount of movement for each hand, measured as the maximum distance between two points in the trajectory of the hand. The movement in the left hand is significantly smaller than that of the right hand in many signs, consistent with the fact that the right hand is the dominant one for all the signers.

\section{Preprocessed version}

\label{sec:preprocessing}

We provide a pre-processed version of the dataset to alleviate the overhead of performing experiments with the data.

\begin{figure}
\centering
\includegraphics[width=0.7\textwidth]{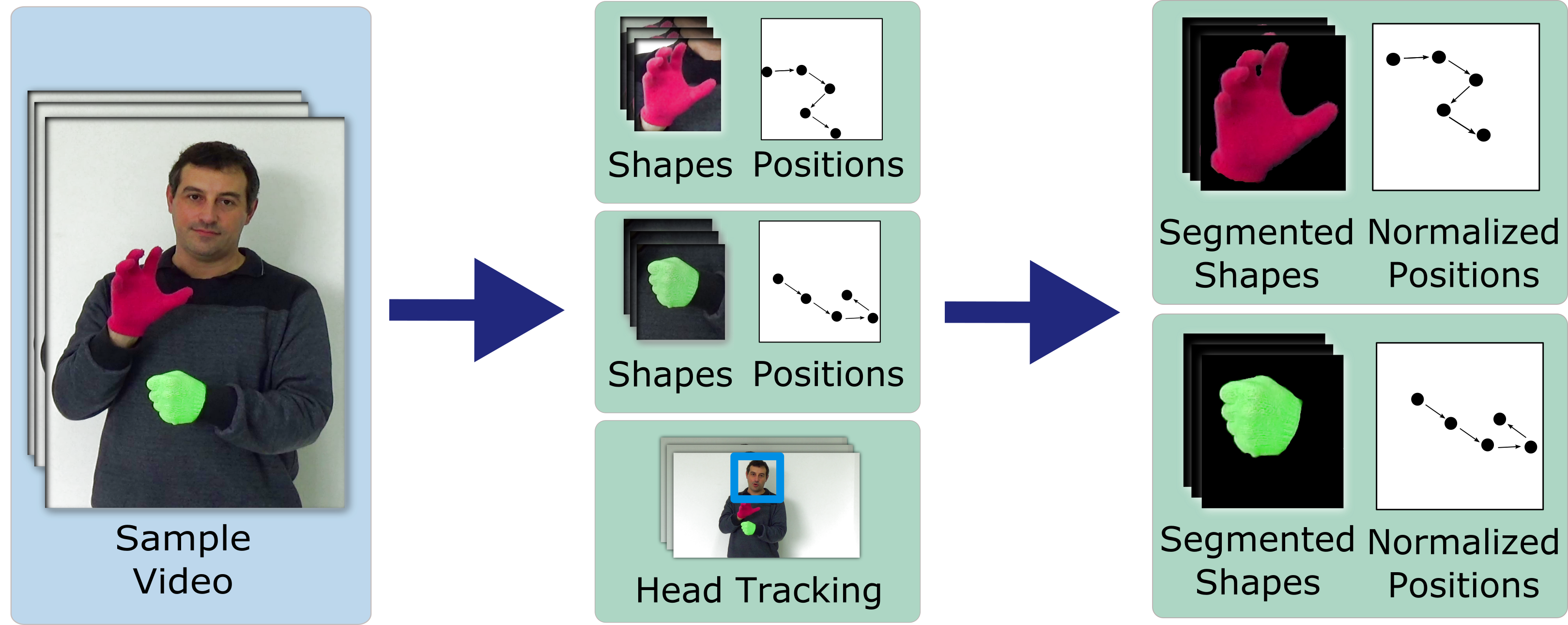}
\caption{Feature extraction steps for the preprocessed version of LSA64. From the sample video, we track the position of both hands and head in all frames. The shapes of each hand are segmented and the positions of them are re-centered with respect to those of the head. }
\label{fig:preprocessing}
\end{figure}

From the dataset we extracted the hand and head positions for each frame, along with images of each hand, segmented and with a black background, as show in Figure \ref{fig:preprocessing}.


The tracking and segmentation of the hands uses the techniques described in \cite{Ronchetti2016}. Additionally, the head of the subject is tracked via the Viola-Jones's face detector \cite{viola2004robust}. The 2D position of each hand is translated so that the head is at the origin. The positions are then normalized by dividing by the arm's length of the subject, measured in centimeters/pixels. In this way, the transformed positions represent distances from the head, in units of centimeters.

The result of this process is a sequence of frame information, where for each frame we calculate the position of both hands, and we extract an image of each hand with the background segmented.

\section{Baseline Experiments}

In this section, we briefly describe the model and results obtained in signer-dependent and independent experiments with the dataset, to establish a baseline performance. The model we used is described fully in \cite{Ronchetti2016b}.

The model we used to get a baseline performance on the dataset classifies the information for each hand separately and then multiplies the probabilities output by the subclassifiers, per class. The model for each hand includes three subclassifiers, each processing position, movement or handshape information.

The movement subclassifier contains one left-to-right Hidden Markov Model (HMM) per class, with skip transitions. All the models have 4 states. The output probabilities are modeled with a Gaussian Mixture Model (GMM) in each state. The models are trained with EM with the trajectories computed in the pre-processed version of the dataset.

The handshape subclassifier also employs HMM-GMMs, but uses as input the output of the static handshape classifier described in \cite{Ronchetti2016}. The static classifier is run on the segmented hand for each frame of the video, and the sequence of probabilities is fed into the handshape HMM-GMM to obtain the probability of each class for the whole sequence of frames.

The position subclassifier models the initial and final positions of the signs of each class with a set of gaussian distributions. There are two gaussian per class, one for the initial and another one for the final position.

For each class, the model outputs the product of the probabilities given by the position, movement and handshape information of both hands. For one-handed gestures, the information of the left hand is ignored, and so the probabilities output by the left hand model are not multiplied.

We performed a subject-dependent classification experiment with the model, using stratified repeated random sub-sampling validation as the cross-validation scheme, with 30 runs and an 80-20 training-test split. For each run, we measured the classification accuracy of the model on the test dataset. The mean accuracy obtained was 95.95\% (standard deviation $\sigma=0.954$).

\section{Conclusion}

We have presented a dataset of signs from the Argentinian Sign Language. To the best of our knowledge, there are currently no research-oriented datasets of this language created or available.

The subjects used colored gloves in the recording of the signs to significantly ease the tracking and segmentation steps. Nonetheless, we also provide a pre-processed version of the dataset to facilitate experimentation and reproducibility.

We have also presented a set of statistics and extra information to characterize the dataset and allow researchers to easily understand its nature. The signs in this dataset possess significant overlap in terms of types of movements, initial and final positions and handshapes, producing non-trivial experiment settings to test new sign language recognition models.

We intend to expand the dataset with both new signs and a set of annotated LSA sentences to provide a complete basic working vocabulary for Argentinian Sign Language.

\bibliographystyle{splncs03}

\bibliography{bib/general,bib/psom,bib/surrey,bib/triesch,bib/midusi}

\begin{thebibliography}{1}
\providecommand{\url}[1]{\texttt{#1}}
\providecommand{\urlprefix}{URL }

\bibitem{Cooper2011a}
Cooper, H., Holt, B., Bowden, R.: Sign language recognition. In: Moeslund,
  T.B., Hilton, A., Kr\"{u}ger, V., Sigal, L. (eds.) Visual Analysis of Humans:
  Looking at People, chap.~27, pp. 539 -- 562. Springer (Oct 2011)

\bibitem{kapuscinski2015recognition}
Kapuscinski, T., Oszust, M., Wysocki, M., Warchol, D.: Recognition of hand
  gestures observed by depth cameras. International Journal of Advanced Robotic
  Systems  12 (2015)

\bibitem{neidle2012challenges}
Neidle, C., Thangali, A., Sclaroff, S.: Challenges in development of the
  american sign language lexicon video dataset (asllvd) corpus. In: Proc. 5th
  Workshop on the Representation and Processing of Sign Languages: Interactions
  between Corpus and Lexicon. Citeseer (2012)

\bibitem{Ronchetti2016b}
Ronchetti, F., Quiroga, F., Estrebou, C., Lanzarini, L., Rosete-Suárez, A.:
  Sign languague recognition without frame-sequencing constraints: A proof of
  concept on the argentinian sign language. IBERAMIA: Iberoamerican Society of
  Artificial Intelligence  (2016)

\bibitem{Ronchetti2016}
Ronchetti, F., Quiroga, F., Lanzarini, L., Estrebou, C.: Handshape recognition
  for argentinian sign language using probsom. Journal of Computer Science and
  Technology  16(1),  1--5 (2016)

\bibitem{Roussos2010a}
Roussos, A., Theodorakis, S., Pitsikalis, V., Maragos, P.: Hand tracking and
  affine shape-appearance handshape sub-units in continuous sign language
  recognition. In: Trends and Topics in Computer Vision - {ECCV} 2010
  Workshops, Heraklion, Crete, Greece, Revised Selected Papers, Part{I}. pp.
  258--272 (2010)

\bibitem{viola2004robust}
Viola, P., Jones, M.J.: Robust real-time face detection. International journal
  of computer vision  57(2),  137--154 (2004)

\bibitem{von2008recent}
Von~Agris, U., Zieren, J., Canzler, U., Bauer, B., Kraiss, K.F.: Recent
  developments in visual sign language recognition. Universal Access in the
  Information Society  6(4),  323--362 (2008)

\bibitem{wang2009real}
Wang, R.Y., Popovi{\'c}, J.: Real-time hand-tracking with a color glove. ACM
  transactions on graphics (TOG)  28(3), ~63 (2009)

\end{thebibliography}

\end{document}